\def\year{2019}\relax
\documentclass[letterpaper]{article} 
\usepackage{aaai19}  
\usepackage{times}  
\usepackage{helvet}  
\usepackage{courier}  
\usepackage{url}  
\usepackage{graphicx}  
\usepackage{latexsym}
\usepackage{amsmath}
\usepackage{multirow}
\usepackage{subcaption}
\begin{document}
%
\title{eRevise: Using Natural Language Processing to Provide Formative Feedback on Text Evidence Usage in Student Writing}
\author{H. Zhang, A. Magooda, D. Litman, R. Correnti, E. Wang, L.C. Matsmura, E. Howe,  R. Quintana\\
Learning Research and Development Center\\
University of Pittsburgh\\
Pittsburgh, Pennsylvania 15260\\
}
\maketitle
\begin{abstract}
Writing a good essay typically involves students  revising an initial paper draft after receiving feedback.  We present eRevise, a web-based writing and revising environment that uses natural language processing features generated for rubric-based essay scoring to trigger formative feedback messages regarding students' use of evidence in response-to-text writing. By helping students  understand the criteria for using text evidence during writing, eRevise empowers students to better revise their paper drafts. In a pilot deployment of eRevise  in 7 classrooms spanning grades 5 and 6, the quality of  text evidence usage in writing improved after students received formative feedback then engaged in paper revision.
\end{abstract}

\section{Introduction}
With benefits such as minimizing human effort and assuring scoring consistency, natural language processing (NLP) has been used to develop many Automatic Essay Scoring (AES) systems that can reliably assess the content, structure, and quality of written prose 
\cite{shermis2003automated,shermis2013handbook}.  However, before providing students with final essay scores,  
engaging students in cycles of essay drafting and revising after feedback is also essential \cite{graham2015based}.  This is because scoring without feedback is typically not enough to guide students on how to revise an essay for further improvement. Standalone AES systems are thus  increasingly being incorporated into Automated Writing Evaluation (AWE) systems~\cite{dikli2006overview,roscoe2014writing,weigle2013english},
which provide students with formative feedback in addition to (or instead of) essay scores.
Formative feedback can guide students during revision in ways that help students compensate for identified essay weaknesses. Although   \citeauthor{foltz2015analysis} \shortcite{foltz2015analysis}  showed that 
student 
writing could improve with revisions based on AWE feedback and \citeauthor{chapelle2015validity} \shortcite{chapelle2015validity} showed that successful revising is related to feedback accuracy, much AWE research remains to be done.


This paper presents the design and first classroom evaluation of eRevise, an AWE system for improving students' ability to use text evidence 
-- a
dimension of writing that is important for college and career readiness.
eRevise has been designed for  students in grades 5-6 taking the Response to Text Assessment (RTA)~\cite{correnti2013assessing}, where students first write an essay in response to a source text and are then assessed using a detailed Evidence scoring rubric.\footnote{Although eRevise currently focuses only on the Evidence rubric, the full RTA as scored by humans comprises five dimensions: Analysis, Evidence, Organization, Academic Style, and MUGS (Mechanics, Usage, Grammar, Spelling).} In particular, eRevise has been developed for deployment in a formative/classroom environment over two class periods (in contrast to a summative/high-stakes usage). Students write their essays in the first period, then revise their essays in the second period after receiving  formative feedback automatically selected based on  first draft AES. 
In contrast to many AES systems that achieve high scoring reliability but do not address construct validity \cite{condon2013large,perelman2012construct}, eRevise uses a rubric-based AES system to ensure that dimensions of the construct are well represented by the indicators used for scoring \cite{loukina2015feature}. This in turn enables the development of an AWE algorithm for converting internal AES data structures into formative feedback messages that  are both tailored to each student's writing needs and aligned to the constructs of the scoring rubric. eRevise is also notable in focusing on evidence usage rather than on surface writing features, and on upper elementary rather than middle or high school students, which makes the application of NLP techniques particularly challenging.

The next two sections describe the eRevise workflow, and the NLP
techniques supporting eRevise's AES and AWE components, respectively. This is followed by a classroom deployment and evaluation section demonstrating the
promise of eRevise in supporting essay
revision.
Our deployment tested whether  eRevise  helped students: 1) improve the overall quality of their drafts when evaluated by human scorers using the RTA evidence rubric, and 2) increase the quantity and relevance/specifity of their text evidence usage
when evaluated using NLP.  Analyses of 143 
essays  created by 5th and 6th grade students from 7 different classes 
support both hypotheses.

\section{System Usage and Architecture}

\begin{figure}[t]
\centering
\includegraphics[height=5cm]{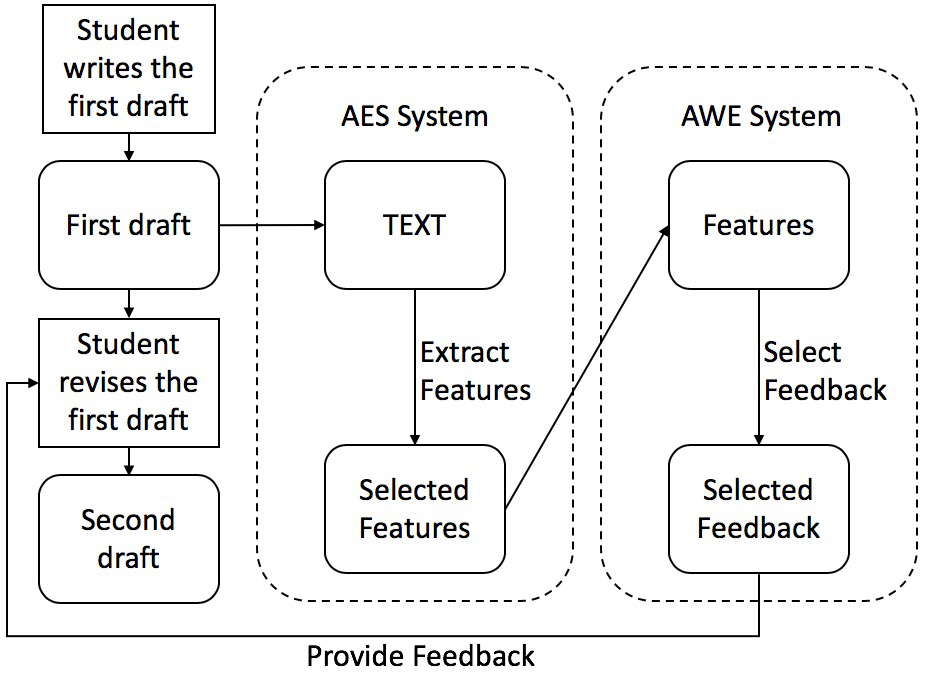}
\caption{The architecture of the eRevise system.}
\label{fig:overview}
\end{figure}




\begin{figure*}[ht]
\centering
\includegraphics[height=7cm]{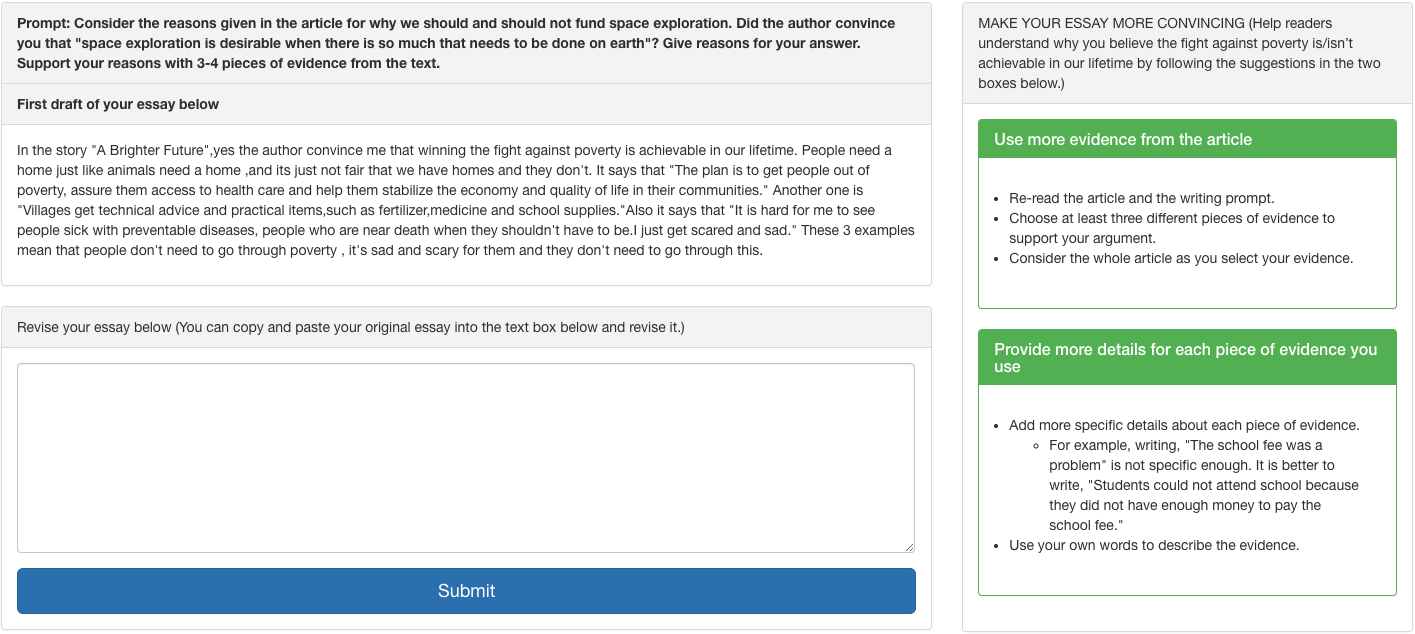}

\caption{A formative feedback screenshot of the eRevise system.}
\label{fig:screenshot}
\end{figure*}

During the first class, a teacher reads an article 
aloud (with students following along on a hardcopy) about 
an effort by the Millenium Villages Project (MVP) to eradicate poverty in a 
Kenyan village.\footnote{While the RTA has three forms (i.e., articles), eRevise currently only supports AES for 
$RTA_{MVP}$.} After the teacher discusses predefined 
vocabulary and asks standardized questions at designated points, there is a prompt at the end of the article which asks students: ``Based on the article, did the author provide a convincing argument that winning the fight against poverty is achievable in our lifetime?  Explain why or why not with 3-4 examples from the text to support your answer.'' At this point students use eRevise to write an essay in response to the prompt. Both the source article and the prompt appear on the screen, with students typing their drafts into a text area below the prompt. 
The purpose of this first usage of eRevise is to electronically collect students' first drafts. 

Figure~\ref{fig:overview} shows the architecture of eRevise. 
After students submit their first drafts, eRevise's AES component uses a previously developed RTA Evidence scoring algorithm~\cite{zhang2017word} to extract features representing the quality of text-based evidence usage  in terms of constructs in the RTA Evidence rubric.  Some of these features (described in the next section) are then passed as input to the AWE system's feedback selection algorithm, which will in turn output a subset of predefined feedback messages that are believed to best address the problems of the first draft based on the features.  These formative feedback messages (although not the AES Evidence scores themselves) will be shown to students during the second class period. 

During the second class period, students login to eRevise  and now revise their first drafts using eRevise. Figure~\ref{fig:screenshot} shows a screenshot illustrating revision guided by formative feedback. 
The left top box shows a student's first draft. This helps students to recall their first drafts and eases revising (e.g., by allowing cutting and pasting). The right-hand side of the screen shows the feedback on the first draft that was automatically selected by the AWE system. The left bottom box shows where students create their second drafts, hopefully guided by the feedback displayed on the right.\footnote{After a student clicks submit, the AES system also scores the revised version of the student's essay. Although eRevise does not share  AES scores with students (due to its focus on formative feedback rather than summative assessment), AES scores are included in  summary reports later shared with teachers.}


\section{Essay Analysis and Feedback Selection}
The ultimate goal of our 
research is to dynamically generate 
formative feedback 
that  
incorporates excerpts from students' essays. However, 
to simplify 
the first version of eRevise, the current AWE system instead dynamically selects two of four pre-defined feedback messages to guide students in revising first drafts. Table~\ref{tab:feedbacks} shows 
these messages,
ordered by a typical progression of evidence usage in student writing development. The messages reflect research on effective feedback and conceptual
frameworks for effective text evidence use~\cite{wang18}, and were created by content experts after an analysis of previously scored RTA essays~\cite{correnti2013assessing,rahimi2014automatic,rahimi2017assessing,zhang2017word}. 
These 4 messages 
were then organized into groups of two  based on the appropriateness of the messages for essays with differing evidence sophistication (messages 1 and 2 for the least sophisticated, messages 2 and 3 for more sophisticated, and messages 3 and 4 for the most sophisticated essays). 
Based on AES feature analysis of evidence usage in the first draft  (and further feature processing, described below), each student thus receives  two  feedback messages based on the group assigned to the essay.

\begin{table*}[ht]
\begin{center}
\small
\begin{tabular}{|p{0.03\linewidth}|p{0.17\linewidth}|p{0.7\linewidth}|}
\hline \bf No. & \bf Name & \bf Feedback  \\ \hline
1 & Use more evidence from the article & 
\textbullet Re-read the article and the writing prompt.

\textbullet Choose at least three different pieces of evidence to support your argument.

\textbullet Consider the whole article as you select your evidence.
\\ \hline
2 & Provide more details for each piece of evidence you use & 
\textbullet Add more specific details about each piece of evidence.

\quad \textendash For example, writing, ``The school fee was a problem'' is not specific enough. It is better to write, "Students could not attend school because they did not have enough money to pay the school fee."

\textbullet Use your own words to describe the evidence.
\\ \hline
3 & Explain the evidence & 
\textbullet Tell your reader why you included each piece of evidence. Explain how the evidence helps to make your point.
\\ \hline
4 & Explain how the evidence connects to the main idea \& elaborate & 
\textbullet Tie the evidence not only to the point you are making within a paragraph, but to your overall argument.

\textbullet Elaborate. Give a detailed and clear explanation of how the evidence supports your argument.
\\
\hline
\end{tabular}
\end{center}
\caption{\label{tab:feedbacks} Four feedback messages predefined by  content experts, based on progression of evidence use.}
\end{table*}

\begin{table*}
\centering
\scalebox{0.6}{
\begin{tabular}{|p{3.75cm}p{5cm}p{5cm}p{5cm}p{5cm}|}
\hline & \bf 1 & \bf 2 & \bf 3 & \bf 4 \\
\hline Number of Pieces of evidence & Features one or no pieces of evidence (NPE) & Features at least 2 pieces of evidence (NPE) & Features at least 3 pieces of evidence (NPE) & Features at least 3 pieces of evidence (NPE) \\ \hline
Relevance of evidence & Selects inappropriate or irrelevant details from the text to support key idea (SPC); references to text feature serious factual errors or omissions & Selects some appropriate and relevant evidence to support key idea, or evidence is provided for some ideas, but not actually the key idea (SPC); evidence may contain a factual error or omission & Selects pieces of evidence from the text that are appropriate and relevant to key idea (SPC) & Selects evidence from the text that clearly and effectively supports key idea \\ \hline
Specificity of evidence & Provides general or cursory evidence from the text (SPC) & Provides general or cursory evidence from the text (SPC) & Provides specific evidence from the text (SPC) & Provides pieces of evidence that are detailed and specific (SPC) \\ \hline
Elaboration of Evidence & Evidence may be listed in a sentence (CON) & Evidence provided may be listed in a sentence, not expanded upon (CON) & Attempts to elaborate upon evidence (CON) & Evidence must be used to support key idea / inference(s) \\ \hline
Plagiarism & Summarize entire text or copies heavily from text (in these cases, the response automatically receives a 1) & & & \\
\hline
\end{tabular}}
\caption{\label{rubricTable} Rubric for scoring the Evidence dimension of RTA. The abbreviations in the parentheses identify features used by the AES system that are aligned with specific assessment criteria \cite{rahimi2017assessing}.} 
\end{table*}

\begin{table}[ht]
\begin{center}
\begin{tabular}{|c|c|c|c|c|c|}
\hline \bf AES Model & \bf QWK  \\ \hline
Rubric & 0.632 \\
SG & 0.653 \\
CO-ATTN & 0.697 \\
\hline
\end{tabular}
\end{center}
\caption{\label{tab:aes_result} Quadratic Weighted Kappa (QWK) of different AES models. The {\it CO-ATTN} model significantly outperforms the  {\it Rubric} and  {\it SG} models, respectively ($p<0.05$).}
\end{table}

\begin{table*}[ht]
\begin{center}
\small
\begin{tabular}{|c|c|c|c|c|c|c|c|c|c|c|c|}
\hline \multirow{3}{*}{First Draft} & \bf Essay & \multicolumn{10}{p{12.5cm}|}{
In the story ``A Brighter Future'',yes {\itshape the author convince me that winning} the fight against poverty is achievable in our lifetime. People need a home just like animals need a {\itshape home ,and its just not fair} that we have homes and they don't. It says that ``The plan is {\itshape to get people out of poverty}, assure them access to health care and help them stabilize {\itshape the economy and quality of life} in their communities.'' Another one is ``{\itshape Villages get technical advice and practical} items,such as fertilizer,medicine and school supplies.''Also it says that ``It is hard {\itshape for me to see people sick with preventable diseases}, people who are near death when they shouldn't have to be.I just get scared and sad.'' These 3 examples mean that people don't need to go through poverty , it's sad and scary for them and they don't need to go through this.}\\
\cline{2-12}
& \multirow{2}{*}{\bf{Features}} & NPE & SPC1 & SPC2 & SPC3 & SPC4 & SCP5 & SCP6 & SCP7 & SPC8 & SPC\_Total\_Merged \\
& & 1 & 1 & 2 & 1 & 3 & 0 & 1 & 1 & 2 & 6 \\
\hline \multirow{3}{*}{Second Draft} & \bf Essay & \multicolumn{10}{p{12.5cm}|}{
In the story ``A Brighter Future'' yes {\itshape the author convince me that winning} the fight against poverty is achievable in our lifetime.\ Yes we need to win the fight against poverty because everybody needs a home, shelter, food,and money. It {\bf \itshape say that ``Their crops were dying because they could not afford} the necessary fertilizer and irrigation''Another one is that ``Its hard {\itshape for me to see people sick with preventable diseases,people who are near} death when they shouldn't have to be.''Also ``.{\bf \itshape Little kids were wrapped in cloth} on their {\bf \itshape mothers backs,or running around in bare} feet and tattered clothing.'' These three examples mean that we need to {\bf \itshape help them have a better life} and a better home than the busty,dirty ground.} \\
\cline{2-12}
& \multirow{2}{*}{\bf{Features}} & NPE & SPC1 & SPC2 & SPC3 & SPC4 & SCP5 & SCP6 & SCP7 & SPC8 & SPC\_Total\_Merged \\
& & 3 & 2 & 1 & 2 & 3 & 2 & 1 & 0 & 1 & 6 \\
\hline
\end{tabular}
\end{center}
\caption{\label{essaytable} Examples of a student's first and second essay drafts, showing the NLP analyses during AES that are needed for AWE. For each essay, the text-based evidence identified during AES  that is used to compute the essay's SPC  values is shown in italics (and in bold if only identifed in the second draft). eRevise would display feedback messages 1 and 2 for the first essay draft.}
\end{table*}

\subsection{AES Feature Extraction}

We have developed several AES systems for RTA assessment \cite{rahimi2017assessing,zhang2017word,zhang2018co}. 
Our first model (denoted by $Rubric$)  \cite{rahimi2017assessing} used NLP to represent an essay in terms of features that largely correspond to cells in the RTA Evidence rubric.  This rubric, as well as the correspondence between the rubric and features that serve as input to the scoring model, are shown in Table \ref{rubricTable}.  A subsequent model (denoted by $SG$) 
 \cite{zhang2017word} introduced skip-gram word embeddings into the feature extraction process, in order to increase robustness by moving from lexical to semantic similarity computation. 
Most recently, \citeauthor{zhang2018co} \shortcite{zhang2018co} developed a neural network model with a co-attention layer (denoted by {\it CO-ATTN}) to  eliminate  human feature engineering.
Table \ref{tab:aes_result} shows  performance figures for each of these AES models when evaluated using cross-validation on a previously collected $RTA_{MVP}$ corpus of 2970 essays. Although 
the neural {\it CO-ATTN} model has the best performance, 
to select formative feedback messages that address essay weaknesses in terms of rubric constructs, a more interpretable representation of the essay is necessary. Therefore,  
$SG$ is the AES system used in eRevise.
In particular, two of the  features used by  $SG$  for score prediction, namely Number of Pieces of Evidence (NPE) and Specificity (SPC), form the basis of eRevise's feedback selection algorithm.\footnote{Although  Concentration (CON) is also aligned with the rubric, the other two features are more aligned with the feedback and they are more consequential for improving evidence usage.}
Table~\ref{essaytable} shows an example first and second draft (with AES Evidence scores of 2 and 3, respectively), along with NPE and SPC values. 

{\bf Number of Pieces of Evidence (NPE)} is an integer encoding the number of evidence topics in the article that are mentioned in the essay. A fixed size sliding window algorithm is used to extract this feature. If a window\footnote{In eRevise, all windows are of size 6,  which optimized AES performance on previously scored training essays.} contains at least two similar words from a manually crafted list of main topics and associated words from the article\footnote{For the RTA article, the 4 topics are Hospital, Malaria, Farming, and School~\cite{rahimi2017assessing}.  Computing similarity with pre-defined topics is typical in content-based AES~\cite{Liu14}.}, the window is determined to contain text-based evidence related to the topic. Word embedding is used to calculate word similarity, with two words considered as similar after thresholding, thus enabling both lexical and semantic matching (e.g., a student's use of ``power'' will match  ``electricity'' in the article). 
In Table~\ref{essaytable}, the NPE features indicate that the student used text evidence from more topics after revision, i.e., AES identifies  one topic (Hospital) in the first draft versus three (Hospital, Farming, and Malaria) in the revised draft - although Malaria is actually a false positive.


{\bf Specificity (SPC)} is a vector of integers that encodes the number of specific article examples mentioned in the essay. The length of this vector is the number of manually crafted categories, which is 8 for the RTA article~\cite{rahimi2017assessing}. A window-based algorithm is again used for feature extraction, now using a different manually crafted list of words associated with examples and categories from the article, where all examples are assigned to different categories. 
For example, 
in the sliding window ``\textit{see people sick with preventable diseases}'', the essay words ``\textit{preventable}'' and ``\textit{diseases}'' match  
the article word list (``\textit{malaria common disease preventable treatable}'') for one of the 6 examples associated with SPC category 4\footnote{This category talks about malaria  before the MVP program.}. Therefore, the algorithm 
increments the value of SPC4 by 1.

\begin{table*}[t]
\begin{center}
\small
\begin{tabular}{|c|c|c|c|c|c|c|c|c|c|c|c|c|c|c|c|}
\hline \bf Feature & \multicolumn{15}{c|}{\bf Value} \\
\hline \bf $NPE$ & 0 & 0 & 0 & 1 & 2 & 3 & 4 & 1 & 1 & 2 & 2 & 3 & 4 & 3 & 4  \\
\bf $SPC_{lmh}$ & L & M & H & L & L & L & L & M & H & H & M & M & M & H & H \\ \hline
\bf Feedback Messages & 1,2 & 1,2& 1,2 & 1,2 & 1,2 & 1,2 & 1,2 & 1,2 & 1,2 & 1,2 & 2,3 & 2,3 & 2,3& 3,4 & 3,4 \\
\hline
\end{tabular}
\end{center}
\caption{\label{feedbackselection} Lookup table for feedback selection.}
\end{table*}

\subsection{AWE Feedback Selection}

Although the SPC values (which count the number of times the student mentions  specific examples from the article)  
were useful for developing the AES system via supervised machine learning, we found them to be less useful for developing a feedback selection algorithm because the count included duplicate cases, and because the use of word-embedding meant false positive examples were identified during AES. 
The AWE system thus calculates a new feature named $SPC\_Total\_Merged$, which is a count of the number of non-duplicate, unique article examples from the SPC feature vector. 
For example, in the sentence ``{\bf for me to }\underline{{\bf see people sick} with preventable diseases}'', bolding shows the first example found by the algorithm (window-size is 6, matched words are ``people'' and ``sick''), while underlining shows the second (matched words are ``preventable'' and ``diseases''). While the SPC feature considers these as 2 examples, $SPC\_Total\_Merged$ considers them as 1 unique example. For the first draft in Table~\ref{essaytable}, the algorithm thus reduces the SPC total
of 11 (from AES, equation below) to a smaller merged total of 6 (for AWE). 

After extracting the above features for our previously collected corpora of scored essays, AWE feedback selection was guided by three assumptions: 1) the NPE feature indicates the breadth of unique topics, 2) the SPC\_Total\_Merged feature indicates the number of unique pieces of evidence the student located and used; and 3) a matrix of these two indicators could match each essay to appropriate feedback. Given we did not need to modify NPE, the following equations were used to calculate $SPC_{AWE}$ for feedback selection.
\[
SPC_{total} = \sum_{i=1}^{N}{SPC_i} \tag{1} \label{eq:1}
\]
$SPC_{total}$, where $N$ is the number of categories in SPC, calculates the total number of matches the computer finds between students' first drafts and examples we are looking for. 

\[
SPC_{important} = \sum_{i=S}^{E}{SPC_i} \tag{2} \label{eq:2}
\]
$SPC_{important}$, where $S$ and $E$ are the start and end indices of important categories, 
calculates  the total number of matched examples from four primary topics for evidence usage (hospital, malaria, farming and school). 
In the $RTA_{MVP}$, content experts identified these categories as primary because they are the topics on which students can provide specific examples of improvement, while other SPC categories refer to more general examples from the article.

\[
DR = \frac{SPC_{total} - SPC\_Total\_Merged}{SPC_{total}} \tag{3} \label{eq:3}
\]
$DR$ calculates the duplication rate of matched examples, by using $SPC\_Total\_Merged$ to calculate the proportion of duplicate evidence from $SPC_{total}$. 

\[
SPC_{AWE} = RND(SPC_{important}*(1 - DR)) \tag{4} \label{eq:4}
\]
$SPC_{AWE}$ adjusts the number of important matched examples by the duplication rate. This produces a new score for generating feedback, representing the number of unique matched examples from four primary topics.
We round the number to get an integer used in the conditional statement below. 

\[
SPC_{lmh}= 
\begin{cases}
    L,& \text{if } SPC_{AWE}<3\\
    M,& \text{if } SPC_{AWE} \geq 3 \text{ and } SPC_{AWE} \leq 5\\
    H,& \text{otherwise}
\end{cases}
\tag{5} \label{eq:5}
\]
$SPC_{lmh}$ is a categorical variable for  $SPC_{AWE}$ that indicates low (L), medium (M), or high (H) values.   

Finally, the AWE system  uses $NPE$ (computed during AES) and $SPC_{lmh}$ to select the two most appropriate feedback messages for the essay based on Table~\ref{feedbackselection}. The content experts used a previously scored corpus \cite{zhang2017word} as development data to manually 
design this table. 

For the first draft in Table~\ref{essaytable}, $NPE=1$, $SPC_{total} = 11$,  $SPC\_Total\_Merged = 6$, 
$SPC_{important} = 6$, $DR = 0.455$. $SPC_{AWE} = 3$, and $SPC_{lmh} = M$. 
Therefore, after consulting Table~\ref{feedbackselection}, eRevise would display feedback messages 1 and 2 
for this essay (as for the essay displayed in Figure~\ref{fig:screenshot}).

In sum, the AWE process results in all students  receiving  two (of four possible) feedback messages that are selected based on the AES feature analysis and are thus targeted to improving the quality of each student's particular essay.  Note that students will receive feedback even when AES predicts a score of 4 for the first draft. In most cases, such students will receive the third and fourth feedback messages focusing on evidence elaboration.


\section{Experimental Deployment and Results}
Our first deployment of eRevise took place  in two public rural parishes 
in
Louisiana. Seven 5th and 6th-grade teachers had all students in one of
their English Language Arts classes write and revise an essay using eRevise.  A total of 
 143 students completed all tasks.
We 
test two hypotheses: 
\begin{description}
\item{H1:} eRevise helps students improve the overall quality of their drafts, as  evaluated by human scorers using the RTA evidence rubric.
\item{H2:} eRevise increases the quantity and relevance/specificity\footnote{Corresponding to row 1 and rows 2-3 of Table~\ref{rubricTable}.} of evidence that students use from the RTA source text,  as evaluated using NLP features.
\end{description}

\begin{figure*}[t]
\centering

\begin{subfigure}[b]{.3\textwidth}
\centering
\includegraphics[width=\linewidth]{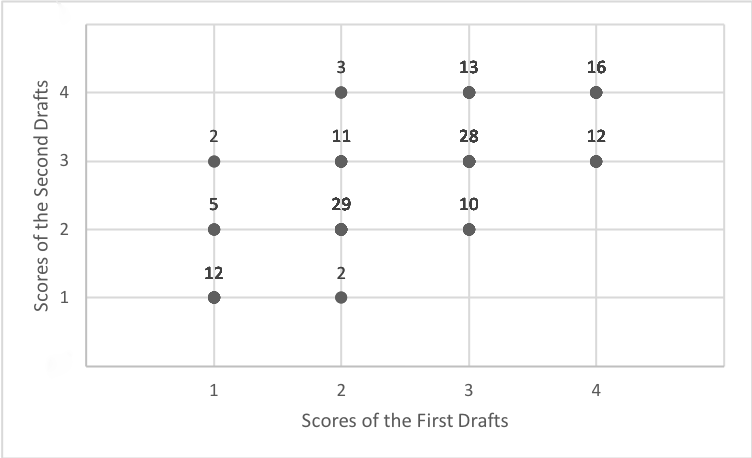}
        \caption{}\label{fig:fig_a}
\end{subfigure}
\begin{subfigure}[b]{.3\textwidth}
\centering
\includegraphics[width=\linewidth]{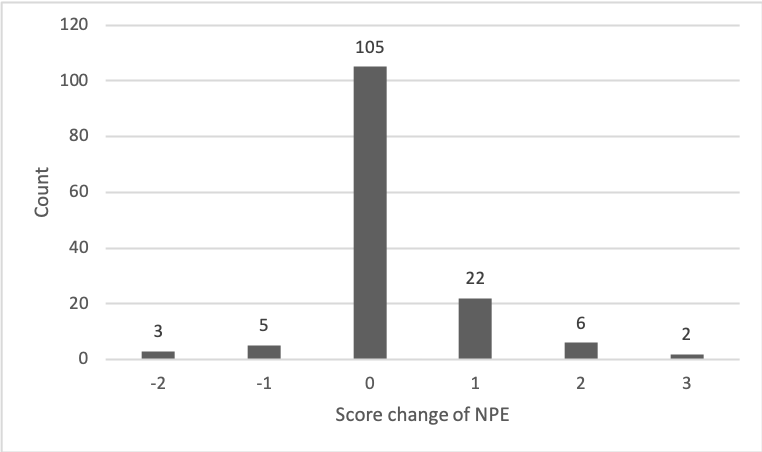}
\caption{}\label{fig:fig_b}
\end{subfigure}
\begin{subfigure}[b]{.3\textwidth}
\centering
\vspace{0pt}
\includegraphics[width=\linewidth]{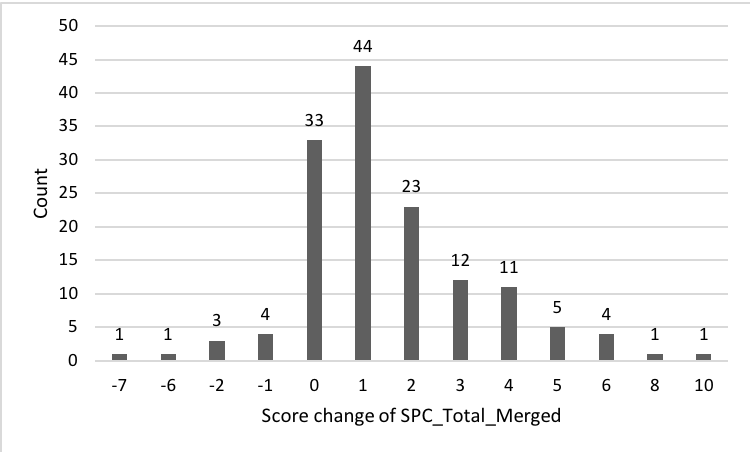}
\caption{}\label{fig:fig_c}
\end{subfigure}

\medskip

\begin{subfigure}[b]{.3\textwidth}
\centering
\includegraphics[width=\linewidth]{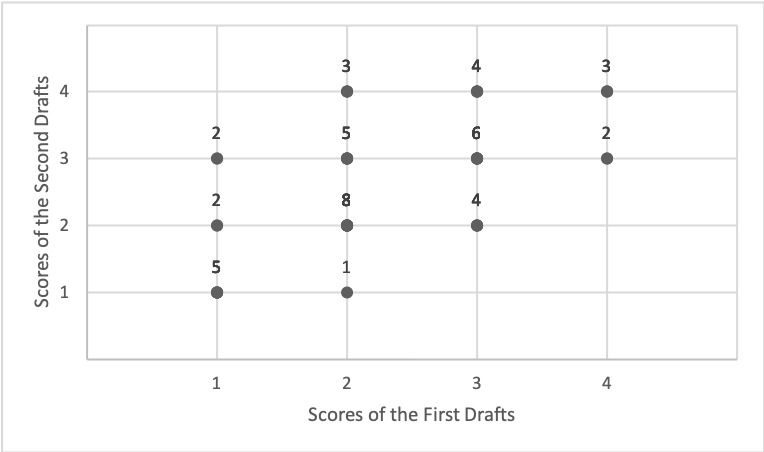}
        \caption{}\label{fig:fig_d}
\end{subfigure}
\begin{subfigure}[b]{.3\textwidth}
\centering
\includegraphics[width=\linewidth]{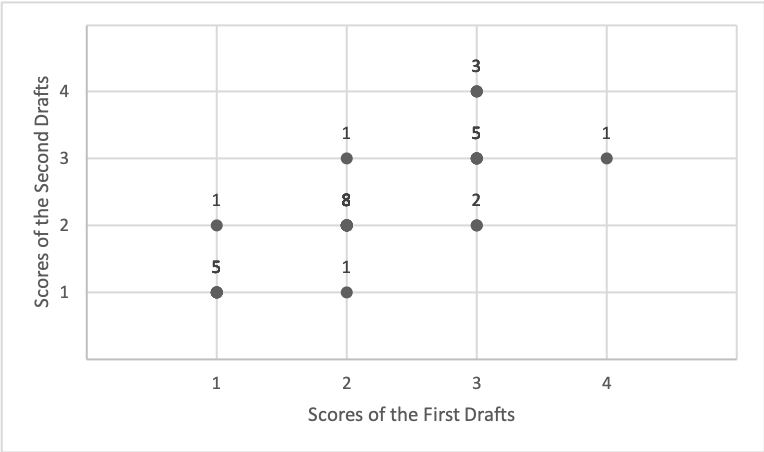}
\caption{}\label{fig:fig_e}
\end{subfigure}
\begin{subfigure}[b]{.3\textwidth}
\centering
\vspace{0pt}
\includegraphics[width=\linewidth]{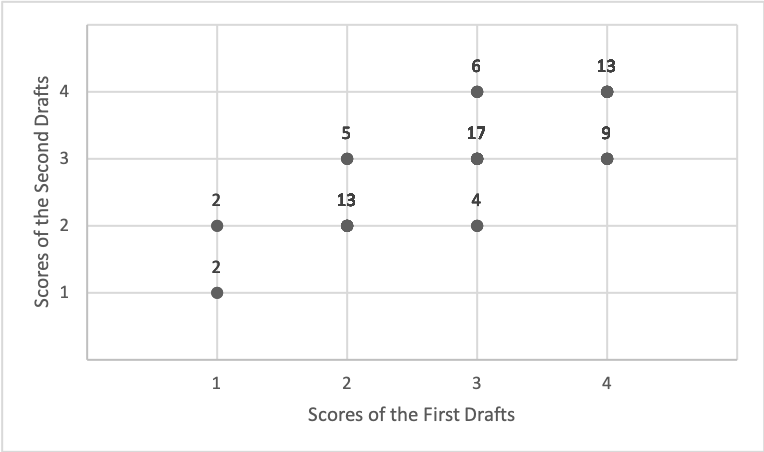}
\caption{}\label{fig:fig_f}
\end{subfigure}

\caption{(\subref{fig:fig_a}) RTA Evidence scores before and after revision. (\subref{fig:fig_b}) Value changes for the NPE feature. (\subref{fig:fig_c}) Value changes for the SPC\_Total\_Merged feature. (\subref{fig:fig_d}) RTA Evidence scores 
for essays receiving feedback messages 1 and 2. (\subref{fig:fig_e}) RTA Evidence scores 
for essays receiving feedback messages 2 and 3. (\subref{fig:fig_f}) RTA Evidence scores 
for essays receiving feedback messages 3 and 4.}
\label{fig:result}

\end{figure*}

The outcome measure for testing H1 is a human-produced RTA Evidence score.   
After the deployment, a trained human grader  used the rubric from Table~\ref{rubricTable} to score all essays, without knowing whether an essay was a first or second draft.
A paired t-test comparing the first and second draft  Evidence scores (n=143) supports H1, as the scores  
improved from first  ($MEAN=2.62$, $SD=0.95$) to second ($MEAN=2.72$, $SD=0.92$) drafts with trending statistical significance ($p \leq 0.08$).  The grader also scored  essays for the other four RTA dimensions (recall footnote 1).   In contrast to Evidence (for which eRevise provided formative feedback to guide revision), there were no  significant or trending score improvements for any of these other RTA dimensions (all $p \geq .29$). 
Finally, the scatter plot in Figure~\ref{fig:fig_a} shows that the overall improvement in the Evidence dimension was observed despite potential ceiling effects:
28 students received the maximum score of 4 on their first drafts, 16 of whom also received the maximum on their second drafts.
The  plot also shows that although the scores increased for 34 students, the scores did not change for the majority of students (and less often even decreased). 

We thus explore the use of more fine-grained outcome measures that  have a stronger relationship to the eRevise feedback that guided student revision. To test H2, we use the $NPE$ and $SPC\_Total\_Merged$ features as automatically computed by eRevise during its deployment to approximate evidence quantity and relevance/specificity, respectively.
Paired t-tests (n=143)
for both 
support H2.  The $NPE$ feature values improved significantly 
($p \leq 0.003$) 
from first ($MEAN=2.61$, $SD=1.27$) to  second draft ($MEAN=2.81$, $SD=1.08$).  The $SPC\_Total\_Merged$ feature values  also improved 
significantly ($p \leq 0.001$) from  first ($MEAN=9.65$, $SD=4.94$) to second drafts ($MEAN=11.15$, $SD=5.39$).
For NPE,
the histogram in Figure~\ref{fig:fig_b} shows that  more students added rather than removed evidence (30 versus 8 students). Although 105 students showed no evidence change, 43 were already at ceiling with NPE values of 4 in the first draft.
For SPC, the histogram in Figure~\ref{fig:fig_c}  shows that a large majority of students (101) increased the number of specific article examples that they incorporated into their essays. 
33 other students showed no change, while only 9 students removed specific examples.

Recall the 16 students in Figure~\ref{fig:fig_a} who were at ceiling when the RTA Evidence score was used as the outcome measure.  By instead using the $SPC\_Total\_Merged$ values as the outcome, these 16 students can now be seen to show improvement  from their first drafts ($MEAN=12.69$, $SD=4.63$) to the second drafts ($MEAN=13.25$, $SD=5.20$), with trending statistical significance ($p \leq 0.095$). 

Finally, 
Figure~\ref{fig:fig_d} shows how evidence scores changed for the 45 essays receiving feedback messages 1 and 2. The evidence score improvements from first ($MEAN=2.33$, $SD=0.93$) to second ($MEAN=2.64$, $SD=0.98$) drafts were statistically significant ($p = 0.02$). Figure~\ref{fig:fig_e} shows the score changes for the 27 essays receiving  feedback messages 2 and 3. The evidence scores only slightly improved from first ($MEAN=2.22$, $SD=0.85$) to second ($MEAN=2.26$, $SD=0.94$) drafts.  Figure~\ref{fig:fig_f} shows that for the 71 essays receiving   messages 3 and 4, the evidence scores were almost the same  from first ($MEAN=2.94$, $SD=0.89$) to second ($MEAN=2.94$, $SD=0.81$) drafts.  These three figures suggest that drafts with the least sophisticated evidence usage had the most room for improvement. It is also interesting to relate these three feedback-based groupings to essay   RTA Evidence scores. 
40.63\% of drafts receiving Evidence scores of 1 or 2 received feedback messages 1 and 2. 62.03\% of drafts  receiving Evidence scores or 3 or 4 received messages 3 and 4. Although only 27 essays received  messages 2 and 3,  71.43\% of these drafts received Evidence scores of 2 or 3.




\section{Current and Future Directions}

We are about to begin the next deployment of eRevise, which will extend our  work in two  ways.  First, to better determine the benefit of using AES to adaptively guide revision, we have added  a control condition  where eRevise will  display the same generic feedback message to all students:  ``MAKE YOUR ESSAY MORE CONVINCING - Help readers understand why you believe the fight against poverty is/isn’t achievable in our lifetime.'' This is in contrast to the existing eRevise adaptive feedback, where students receive different messages based on AES.  
Second,  students will  use eRevise for two different forms of the RTA (i.e., $RTA_{space}$ in addition to $RTA_{MVP}$). While we have already trained a {\it SG}  model for scoring $RTA_{space}$~\cite{zhang2017word},
 Table~\ref{feedbackselection} needs to be verified to ensure the validity of our feedback selection algorithm.  We are also exploring  adding {\it CO-ATTN}  scores to the lookup table.

For the longer term, we  plan to extend our  research in  other ways. To score a new RTA form, human effort is currently necessary to define topical components, e.g., creating a list of topics and a list of examples for scoring $RTA_{space}$. 
While we have developed  pilot data-driven methods that can extract such topical components automatically~\cite{rahimi2016automatically}, our methods need to be improved so that  they do not  degrade {\it SG} model performance. 
eRevise will also be enhanced to provide feedback for Organization, a second substantive RTA writing dimension for which we already have a pilot AES ~\cite{rahimi2017assessing}. 
We also plan to move from feedback selection to more personalized feedback generation, and to create a teacher dashboard which can automatically generate summaries such as Figure~\ref{fig:fig_a}. 
Finally, since  eRevise's feedback
encourages students to add more concrete examples
from the article, some students may simply
copy and paste examples rather than use their
own words as the feedback suggests. While the human RTA
rubric (last row in Table~\ref{rubricTable}) addresses  plagiarism,
eRevise currently does not. We thus plan to incorporate
the detection of different types of adversarial essays into
AES.

\section{Conclusions}
eRevise is an AWE system for text evidence usage that uses NLP features produced by a rubric-based AES system to automatically select formative feedback messages most appropriate to a student's needs.  By increasing  access to  feedback on a substantive and important writing dimension, eRevise has the potential to reduce demands on teachers and to build students' knowledge of effective text evidence usage. 
We first described how eRevise 
uses NLP techniques to evaluate draft essays and to select appropriate formative feedback messages to guide later revision. 
Experimental results from a first deployment in 7 classrooms showed that  eRevise  helped students improve their text evidence usage after receiving  formative feedback and engaging in essay revision.  

\section{Acknowledgments}

The research reported here was supported, in whole or in part, by the Institute of Education Sciences, U.S. Department of Education, through Grant R305A160245 to the University of Pittsburgh. The opinions expressed are those of the authors and do not represent the views of the Institute or the U.S. Department of Education.

\bibliographystyle{aaai}
\bibliography{aaai19}

\end{document}